# One-step regression and classification with crosspoint resistive memory arrays

**Authors:** Zhong Sun, Giacomo Pedretti, Alessandro Bricalli, Daniele Ielmini*

**Affiliations:** Dipartimento di Elettronica, Informazione e Bioingegneria, Politecnico di Milano and IU.NET, Piazza L. da Vinci 32 – 20133 Milano, Italy.

*Correspondence to: daniele.ielmini@polimi.it

**One Sentence Summary:** Machine learning algorithms such as linear regression and logistic regression are trained in one step with crosspoint resistive memory arrays.

**Abstract:** Machine learning has been getting a large attention in the recent years, as a tool to process big data generated by ubiquitous sensors in our daily life. High speed, low energy computing machines are in demand to enable real-time artificial intelligence at the edge, *i.e.*, without the support of a remote frame server in the cloud. Such requirements challenge the complementary metal-oxide-semiconductor (CMOS) technology, which is limited by the Moore's law approaching its end and the communication bottleneck in conventional computing architecture. Novel computing concepts, architectures and devices are thus strongly needed to accelerate data-intensive applications. Here we show a crosspoint resistive memory circuit with feedback configuration can execute linear regression and logistic regression in just one step by computing the pseudoinverse matrix of the data within the memory. The most elementary learning operation, that is the regression of a sequence of data and the classification of a set of data, can thus be executed in one single computational step by the novel technology. One-step learning is further supported by simulations of the prediction of the cost of a house in Boston and the training of a 2-layer neural network for MNIST digit recognition. The results are all obtained in one computational step, thanks to the physical, parallel, and analog computing within the crosspoint array.

# INTRODUCTION

Resistive memories, also known as memristors (*1*), including resistive switching memory (RRAM) and phase change memory (PCM), are emerging as a novel technology for high density storage (*2,3*), neuromorphic hardware (*4,5*) and stochastic security primitives, such as random number generators (*6,7*). Thanks to their ability to store analog values and to their excellent programming speed, resistive memories have also been demonstrated for executing in-memory computing (*8-17*), which eliminates the data transfer between the memory and the processing unit to improve the time and energy efficiency of computation. With a crosspoint architecture, resistive memories can be naturally utilized to perform matrix-vector multiplication (MVM), by exploiting fundamental physical laws such as the Ohm's law and the Kirchhoff's law of electric circuits (*8*). Crosspoint MVM has been shown to accelerate various data-intensive tasks, such as training and inference of deep neural networks (*11-14*), signal and image processing (*15*), and the iterative solution of a system of linear equations (*16*) or a differential equation (*17*). With a feedback circuit configuration, the crosspoint array has been shown to solve systems of linear equations and calculate matrix eigenvectors in one step (*18*). Such a low computational complexity is attributed to the massive parallelism within the crosspoint array, and to the analog storage and computation with physical MVM. Here, we show that a crosspoint resistive memory circuit with feedback configuration is able to accelerate fundamental learning functions, such as predicting the next point of a sequence by linear regression, or attributing a new input to either one of two classes of objects by logistic regression. These operations are completed in just one step in the circuit, in contrast to the iterative algorithms running on conventional digital computers, which approach the solution with a polynomial time complexity.

# RESULTS

## Linear regression in one step

Linear regression is a fundamental machine learning (ML) model for regressive and predictive analysis in various disciplines, such as biology, social science, economics and management (*19-21*). Logistic regression, instead, is a typical tool for classification tasks (*22*), *e.g.*, acting as the last classification layer in a deep neural network (*23,24*). Due to their simplicity, interpretability and well-known properties, linear and logistic regression stand out as the most popular ML algorithms across many fields (*25*). A linear regression model is described by an overdetermined linear system given by:

$$Xw = y, \qquad (1)$$

where $X$ is an $N \times M$ matrix ($N > M$), $y$ is a known vector with a size of $N \times 1$, and $w$ is the unknown weight vector ($M \times 1$) to be solved. As the problem is overdetermined, there is typically no exact solution $w$ to Eq. (1). The best solution of Eq. (1) can be obtained by the least squares error (LSE) approach, which minimizes the norm of error $\varepsilon = Xw - y$, namely $\|\varepsilon\| = \|Xw - y\|_2$ where $\|\cdot\|_2$ is the Euclidean norm. The vector $w$ minimizing $\|\varepsilon\|$ is obtained by the pseudoinverse (*21,23,24*) [or Moore-Penrose inverse (*26*)] $X^+$, given by:

$$w = X^+ y = (X^T X)^{-1} X^T y, \qquad (2)$$

where $X^T$ is the transpose of matrix $X$.

To obtain the solution in Eq. (2), we propose a crosspoint resistive memory circuit in Fig. 1A, where the matrix $X$ is mapped by the conductance matrix $G_X$ in a pair of crosspoint arrays of analog resistive memories, the vector $y$ corresponds to the opposite of the input current vector $i = [I_1; I_2; ...; I_N]$, and $w$ is represented by the output voltage vector $v = [V_1; V_2; ...; V_M]$ ($M = 2$ in Fig. 1A).

For a practical demonstration of this concept, we adopted arrays of RRAM devices composed of a HfO$_2$ dielectric layer sandwiched between a Ti top electrode and a C bottom electrode (*27*). This type of RRAM device can be programmed to any arbitrary analog conductance within a certain range, thus allowing to represent the matrix elements $X_{ij}$ of the matrix $X$ with sufficient accuracy (*18*). Representative analog conductance levels were programmed by controlling the compliance current during the set transition, as shown in Fig. S1. The crosspoint arrays are connected within a nested feedback loop (*28*) by $N$ operational amplifiers (OAs) from the left array to the right array, and $M$ OAs from the right array to the left array. Briefly, the first set of OAs gives a negative transfer of the current, while the second one gives a positive transfer, resulting in an overall negative feedback, hence stable operation of the circuit with virtual ground inputs of all OAs. A detailed analysis of the circuit stability is reported in Supplementary text 1.

According to Ohm's law and Kirchhoff's law in Fig. 1A, the input currents at the OAs from the left crosspoint array are $G_X v + i$, thus the output voltages applied to the right crosspoint array are $v_r = -\frac{(G_X v + i)}{G_{TI}}$, where $G_{TI}$ is the feedback conductance of transimpedance amplifiers (TIAs). The right crosspoint array operates another MVM between the voltage vector $v_r$ and the transpose conductance matrix $G_X^T$, resulting in a current vector $-G_X^T \frac{(G_X v + i)}{G_{TI}}$, which is forced to zero at the input nodes of the second set of OAs, namely:

$$G_X^T(G_X v + i) = 0. \qquad (3)$$

The steady-state voltages $v$ at the left array are thus given by:

$$v = -(G_X^T G_X)^{-1} G_X^T i, \qquad (4)$$

which is Eq. (2) with $G_X$, $i$ and $v$ representing $X$, $-y$ and $w$, respectively. The crosspoint array circuit of Fig. 1A thus solves the linear regression problem in just one step.

The circuit of Fig. 1A was implemented in hardware using RRAM devices arranged within a crosspoint architecture on a printed circuit board (see Materials and Methods with Fig. S2). As a basic model, we considered the simple linear regression of points $(x_i, y_i)$, where $i = 1, 2, \ldots, N$, to be fitted by a linear model $w_0 + w_1 x_i = y_i$, where $w_0$ and $w_1$ are the intercept with axis $y$ and the slope, respectively, of the best fitting line. To solve this problem in hardware, we encoded the matrix $X$:

$$X = \begin{bmatrix} 1 & x_1 \\ 1 & x_2 \\ \vdots & \vdots \\ 1 & x_N \end{bmatrix}, \qquad (5)$$

in the crosspoint arrays. A column of discrete resistors with $G = 100$ $\mu$S was used to represent the first column of $X$ in Eq. (5), which is identically equal to 1. The second columns of both arrays were implemented with reconfigurable RRAM devices. A total number of $N = 6$ data points was considered, with each $x_i$ implemented as a RRAM conductance with unit 100 $\mu$S. The unit of conductance was chosen according to the range of linear conduction of the device (*18*), thus ensuring a good computation accuracy. Other aspects such as the current limit of the OAs and the power consumption should also be considered to select the best memory devices in the circuit. Although the conductance values in the two crosspoint arrays should be identical, some mismatch can be tolerated for practical implementations (Supplementary text 2). A program/verify technique was used to minimize the relative error (less than 5%) between the values of $X$ in the two crosspoint arrays (Fig. S3). The data ordinates $-y$ were instead applied as input currents. The input currents should be kept relatively small so that the resulting output signal is low enough to prevent disturbance of the device states in the stored matrix.

Given the matrix $X$ stored in the crosspoint arrays and an input current vector $y$, the corresponding linear system was then solved by the circuit in one step. Fig. 1B shows the resulting dataset for an input current vector $i = [0.3; 0.4; 0.4; 0.5; 0.5; 0.6]I_0$ with $I_0 = 100$ $\mu$A to align with the conductance transformation unit. Fig. 1B also shows the regression line, obtained by the circuit output voltages representing weights $w_0$ and $w_1$. The comparison with the analytical regression line shows a relative error of -4.86% and 0.82% for $w_0$ and $w_1$, respectively. The simulated transient behavior of the circuit is shown in Fig. S4, evidencing that the linear regression weights are computed within about 1 $\mu$s. By changing the input vector, a different linear system was formed and solved by the circuit, as shown in Fig. 1C for $i = [0.3; 0.3; 0.5; 0.4; 0.5; 0.7]I_0$. The result evidences that a more scattered dataset can also be correctly fitted by the circuit.

The crosspoint circuit also naturally yields the prediction of the value $y^*$ in respondence of a new point at position $x^*$. This is obtained by adding an extra row in the left crosspoint, where an additional RRAM element is used to implement the new coordinate $x^*$ (Fig. S5). The results are shown in Fig. 1C, indicating a prediction by the circuit which is only 1% smaller compared to the analytical prediction. Fig. S6 reports more linear regression and prediction results of various datasets. Linear regression with 2 independent variables was also demonstrated by a crosspoint array of 3 columns, with results shown in Fig. S7. These results support the crosspoint circuit for the solution of linear regression models in various dimensions. The linear regression concept can also be extended to nonlinear regression models, *e.g.*, polynomial regression (*29*), to better fit a dataset and thus make better predictions. By loading the polynomial terms in crosspoint arrays, the circuit can also realize polynomial regression in one step (Supplementary text 3 with Fig. S8).

**Logistic regression**

Logistic regression is a binary model that is extensively used for object classification and pattern recognition. Different from linear regression which is a fully linear model, logistic regression also includes a nonlinear sigmoid function to generate the binary output. A logistic regression model can be viewed as the single-layer feed-forward neural network in Fig. 2A. Here, the weighted summation vector $s$ of input signals to the nonlinear neuron is given by:

$$s = Xw. \qquad (6)$$

where $X$ is the matrix containing the independent-variable values of all samples, and $w$ is the vector of the synaptic weights. The neuron outputs are thus simply given by the vector $y = f(s)$, where $f$ is the nonlinear function of the neuron. To compute the weights of a logistic regression model with a sample matrix $X$ and a label vector $y$, the logit transformation can be first executed (*30*). By applying the inverse of sigmoid function, the label vector $y$ is converted to a summation vector $s$, namely $s = f^{-1}(y)$. As a result, the logistic regression is reduced to a linear regression problem, where the weights can be obtained in one step by the pseudoinverse concept:

$$w = X^+ s. \qquad (7)$$

For simplicity, we assumed that the nonlinear neuron function is instead a step function and that the summation vector $s$ in Fig. 2A is binarized according to:

$$s_i = \begin{cases} a, if\ y_i = 1 \\ -a, if\ y_i = 0 \end{cases}, \qquad (8)$$

where $a$ is a positive constant for regulating the output voltage in the circuit. After this transformation, the weights can be computed directly with the pseudoinverse circuit of Fig. 1A.

Fig. 2B shows a set of 6 data points with coordinates $(x_1, x_2)$ divided into two classes, namely $y = 0$ (open) and $y = 1$ (full). Fig. 2C shows the matrix $X$ where the first column is equal to $\mathbf{1}$, while the other columns represent the coordinates $x_1$ and $x_2$ of the dataset. The sample matrix $X$ was mapped in the two crosspoint arrays of Fig. 1A, and input current was applied to each row to represent $s$ with $a = 0.2$, according to Eq. (8). The circuit schematic is reported in Fig. S9 together with experimental results and relative errors of logistic regression. The simulated transient behavior of the circuit is shown in Fig. S10, with a computing time around 0.6 $\mu$s. The output voltage yields the weights $w = [w_0; w_1; w_2]$ with $s = w_0 + w_1 x_1 + w_2 x_2 = 0$ representing the decision boundary for classification, where $s \geq 0$ indicates the domain of class '1' and $s < 0$ the domain of class '0'. This is shown as a line in Fig. 2B, displaying a tight agreement with the analytical solution. The crosspoint circuit enables a one-step solution of logistic regression with datasets of various dimensionalities and sizes accommodated by the crosspoint arrays. Similar to linear regression, the circuit can also provide one-step classification of any new (unlabeled) point, which is stored in a grounded additional row of the left crosspoint array. The current flowing in the row yields the class of the new data point. Though here we consider two cases containing only positive independent-variable values for linear/logistic regression, datasets containing negative values can also be addressed by simply translating the entire data to be positive, as explained in Supplementary Fig. S11.

**Linear regression of Boston housing dataset**

While the circuit capability has been demonstrated in experiment for small models, the matrix size is an obvious concern that needs to be addressed for real-world applications. To study the circuit scalability, we considered a large dataset, namely the Boston housing price list for linear regression (*31*,*32*). The dataset collects 13 attributes and the prices of 506 houses, 333 of which are used for training the linear regression model while the rest are used for testing the model. The attributes are summarized in the Supplementary text 4. We performed linear regression with the training set to compute the weights with the crosspoint circuit, applied the regression model to predict house prices of the test set.

Fig. 3A shows the matrix *X* for the training set, including a first column of 1, and the other columns recording the 13 attributes, and the input vector *y*, representing the corresponding prices. The matrix *X* was rescaled to make the conductance values in crosspoint arrays uniform, and the vector *y* was also scaled down to prevent excessive output voltage *w* (see Fig. S12). We simulated the linear regression circuit with SPICE (Simulation Program with Integrated Circuit Emphasis, see Materials and Methods), where the RRAM devices were assumed to accurately map the matrix values within 8-bit precision. Fig. 3B shows the calculated *w* obtained from the output voltage in the simulated circuit, with the relative errors remaining within ±1%, thus demonstrating the good accuracy and scalability of the circuit.

Fig. 3C shows the obtained regression results compared with the real house prices of the training set. A standard deviation $\sigma_P$ of 4,733\$ is obtained from SPICE simulations, which is in line with the analytical solution $\sigma_P' = 4,732\$$. Fig. 3D shows the predicted prices of the test set compared to the real prices. The standard deviation from the circuit simulation is $\sigma_P = 4,779\$$, in good agreement with the analytical results $\sigma_P' = 4,769\$$. The resulting standard deviation is only slightly larger than the training set, which supports the ability for generalization of the model. One-step price prediction for test samples is possible by storing the unlabeled attributes in additional rows of the left crosspoint array and measuring the corresponding current, as indicated in Fig. S13.

**2-layer neural network training**

Logistic regression is widely employed in the last fully-connected classification layer in deep neural networks. The crosspoint circuit thus provides a hardware acceleration of the computation of the last-



layer weights for training of a neural network. To test the crosspoint circuit as an accelerator for training neural networks, we considered the 2-layer perceptron in Fig. 4A, where the first-layer weights are set randomly (*33,34*), while the second-layer weights can be obtained by the pseudoinverse method in the crosspoint circuit. Note that a standard technique to train a 2-layer neural network is the backpropagation algorithm which reduces the squares error iteratively (*35*). In contrast to iterative backpropagation, the pseudoinverse approach can reach the LSE solution in just one step, thus providing a fast and energy-efficient acceleration of network training.

As a case study for neural network training, we adopted the MNIST dataset (*36*). To reduce the circuit size in the simulations, we used only 3,000 out of 50,000 samples to train the neural network. Also, to provide an efficient fan-out (for instance 4) for the first layer (*34*), the image size was down-sampled to be 14×14, resulting in a network of 196 input neurons, 784 hidden neurons and 10 output neurons for the classifications of the digits from 0 to 9. The training matrix $T$ is with a size of 3,000×196, and the first-layer weights $W^{(1)}$ were randomly generated in the range between -0.5 and 0.5 with a uniform distribution (Fig. S14). The matrix $X$ can thus be obtained by:

$$X = f(TW^{(1)}), \qquad (9)$$

while the weights of the second layer $W^{(2)}$ can be obtained by the pseudoinverse model of Eq. (2), with $Y$ containing all the known labels of training samples transformed according to Eq. (8) with $a = 0.05$. For each training sample, the neuron corresponding to the digit is labelled 1, while the other 9 neurons are 0. Note that the matrix $X$ results from the output of a sigmoid function of hidden neuron, thus is restricted in the range between 0 and 1 (Fig. S15).

Fig. 4B shows the second-layer weights $W^{(2)}$ obtained by the simulation of the crosspoint circuit, where $X$ was stored in the RRAM devices and each column of matrix $Y$ was applied as input current. The weights were obtained in ten steps, one for each classification output (from digit '0' to digit '9'). With the computed weights $W^{(2)}$, the network can recognize 500 handwritten digits with accuracy of 94.2%,



which is identical to the analytical pseudoinverse solution. For the whole test set (10,000 digits), the recognition accuracy is 92.15% using the simulated $W^{(2)}$, compared to 92.14% using the analytical solution. The crosspoint array can thus be used to accelerate the training of typical neural networks with ideal accuracy. The computed weights can then be stored in one or more open-loop crosspoint array for accelerating the neural network in the inference mode by exploiting in-memory MVM (Fig. S16) (*8,11,12*).

Fig. 4B also shows the LSEs obtained from both the circuit simulation and analytical study. Note that the LSEs are different among the 10 digits, due to the dependence of LSEs on weight values (Supplementary text 5). Fig. 4C shows the simulated weights as a function of the analytical values for each output neuron, showing a good consistency except for the bias weight $w_0$. The bias acts as a regulator to the summation of an output neuron, thus the deviated bias weight guarantees that the simulated LSE is close to the analytical one in Fig. 4B. It should be noted that, although a random $W^{(1)}$ was assumed in this study, $W^{(1)}$ can be further optimized by gradient descent methods (*37*) to improve the accuracy. The same approach might be applied to pre-trained deep networks by the concept of transfer learning (*38*), thus enabling the one-step training capability for a generalized range of learning tasks.

**DISCUSSION**

Although the crosspoint circuit is inherently accurate and scalable, the imperfections of RRAM devices such as conductance discretization and stochastic variation (*18*) might affect the solution. To study the impact of these issues on the solution accuracy, we assumed a RRAM model with 32 discrete conductance levels, including 31 uniformly-spaced levels and one deep high resistance state (HRS), which is achievable in many resistive memory devices (*39-41*). The ratio between the maximum conductance $G_{max}$ and the minimum conductance $G_{min}$ is assumed to be $\frac{G_{max}}{G_{min}} = 10^3$, in line with previous reports (*42,43*). To describe conductance variations, we assumed a standard deviation $\sigma = \Delta G/6$, $\Delta G/4$ or $\Delta G/2$, where $\Delta G$ is the nominal difference between 2 adjacent conductance levels. The



simulation results for the Boston housing benchmark (Fig. S17) shows that the resulting regression and prediction remains accurate for all cases. For the worst case ($\sigma = \Delta G/2$), the standard deviation $\sigma_P$ of training set is equal to 4756$, compared to the ideal result of 4732$. The $\sigma_P$ of test set for prediction is even closer to the ideal one, namely 4765$ compared to 4769$. These results highlight the suitability of the crosspoint resistive memory circuit for machine learning tasks, where the device variations can be tolerated for regression, prediction and classification.

Another concern for large scale circuits is the parasitic wire resistance. To study its impact on the accuracy of linear regression for Boston housing dataset, we adopted interconnect parameters at 65 nm technology obtained from the ITRS table (*44*), together with the RRAM model. The results in Fig. S18 show an increased $\sigma_P$ for both regression and prediction, with the latter being more insignificant, which is consistent with the impact of device variation. Specifically, the $\sigma_P$ of prediction becomes merely 4809$, compared to the ideal 4769$, thus supporting the robustness of the linear regression circuit for predictive analysis.

The circuit stability analysis in Supplementary text 1 reveals that the poles of the system all lie in the left half plane, thus the circuit is stable and the computing time is limited by the bandwidth corresponding to the first pole, which is the minimal eigenvalue (or real part of eigenvalue) $\lambda_{min}$ (absolute value) of a quadratic eigenvalue problem (*45*). As $\lambda_{min}$ becomes larger, the computation of the circuit gets faster, with no direct dependence on the size of the dataset. To support this scaling property of the circuit speed, we have simulated the transient dynamics of linear regression of the Boston housing dataset and its subsets for increasing size of the training samples (Fig. S19). The results show that the computing time may even decrease as the number of samples increases, which can be explained by the different $\lambda_{min}$ of the datasets (Fig. S20). These results evidence that the time complexity of the crosspoint circuit for linear regression substantially differs from its counterparts of classical digital algorithms, with a potential of approaching size-independent time complexity to significantly speed up large scale ML problems. Note that, as the circuit size increases, a larger current is also required to sustain the circuit operation, which



might be limited by the capability of the OAs. To control the maximum current consumption in the circuit, the memory element should be carefully optimized by materials and device engineering (*46*) or by novel device concepts such as electrochemical transistor (*47,48*), to provide a low-conductance implementation. The impact of device variations and the energy efficiency of the circuit are studied for the 2-layer neural network for MNIST dataset training (Supplementary text 6 with Fig. S21). The results support the robustness of the circuit against device variations for classification applications, and an energy efficiency of 45.3 tera-operations per second per Watt (TOPS/W), which is 19.7 and 6.5 times better than the Google's tensor processing unit (TPU) (*49*) and a highly optimized application-specific integrated circuit (ASIC) system (*50*), respectively.

In conclusion, the crosspoint circuit has been shown to provide a one-step solution to linear regression and logistic regression, which is demonstrated in experiments with RRAM devices. The one-step learning capability relies on the high parallelism of analog computing by physical Ohm's law and Kirchhoff's law within the circuit, as well as physical iteration within the nested feedback architecture. Scalability of the crosspoint computing is demonstrated with large problems, such as the Boston housing dataset and the MNIST dataset. The results evidence that in-memory computing is significantly promising for accelerating machine learning tasks with high latency/energy performance, in a wide range of data-intensive applications.



## Materials and Methods

### RRAM device fabrication

The RRAM devices in this work use a 5-nm $HfO_2$ thin film as the dielectric layer, which was deposited by e-beam evaporation on a confined graphitic C bottom electrode (BE). Without breaking the vacuum, a Ti layer was deposited on top of the $HfO_2$ layer as top electrode (TE). The forming process was operated by applying a DC voltage sweep from 0 to 5 V, where with the voltage was applied to the TE and the BE was grounded. After the forming process, the set and reset transitions take place under positive and negative voltages applied to the TE, respectively.

### Circuit experiment

For all the experiments, the devices were arranged in the crosspoint configuration on a custom Printed Circuit Board (PCB, see Fig. S2), and an Agilent B2902A Precision Source/Measure Unit was employed to program the devices to different conductance states.

Linear and logistic regression pseudoinverse experiments were carried out on a custom PCB with operational amplifiers (OAs) of model AD823 (Analog Devices) for the Negative Feedback Amplifiers (NFA) and OP2177 (Analog Devices) for Positive Feedback Amplifiers (PFA). RRAM devices of left matrix were connected with the BE to the NFAs' inverting-input nodes and with the TE to the PFAs' output terminals. RRAM devices of right matrix were connected with the BE to the PFAs' inverting-input nodes and with the TE to the NFAs' output terminals. A BAS40-04 diode is connected between every amplifier and ground, to limit the voltages within $\pm 0.7$ V, avoiding conductance changes of RRAM devices.

All the input signals were given by a 4-channels arbitrary waveform generator (Aim-TTi TGA12104) and applied to fixed input resistances, which were connected between the input and the NFAs' inverting-input nodes. The PFAs' output voltages were monitored by an oscilloscope (LeCroy Wavesurfer 3024). The board was powered by a BK Precision 1761 DC power supply.

### SPICE simulation



Simulations of the crosspoint circuit for Boston housing case and MNIST training were carried out using LTSPICE (https://www.linear.com/solutions/1066). Linear resistors with defined conductance values were used to map a matrix in the crosspoint arrays. A universal op-amp model was used for all operational amplifiers, while positive and negative feedback amplifiers have different parameters.

**Acknowledgments:** This article has received funding from the European Research Council (ERC) under the European Union's Horizon 2020 research and innovation programme (grant agreement No 648635). This work was partially performed at Polifab, the micro- and nanofabrication facility of Politecnico di Milano. **Author contributions:** Z.S. conceived the idea and designed the circuit. G.P. designed the printed circuit board. G.P. and Z.S. conducted the experiments. A.B. fabricated the devices. All the authors discussed the experimental and simulation results. Z.S. and D.I. wrote the manuscript with input from all authors. D.I. supervised the research. **Competing interests:** The authors declare that they have no competing interests. **Data and materials availability:** All data needed to evaluate the conclusions in the paper are present in the paper and/or the Supplementary Materials. Additional data available from authors upon request.



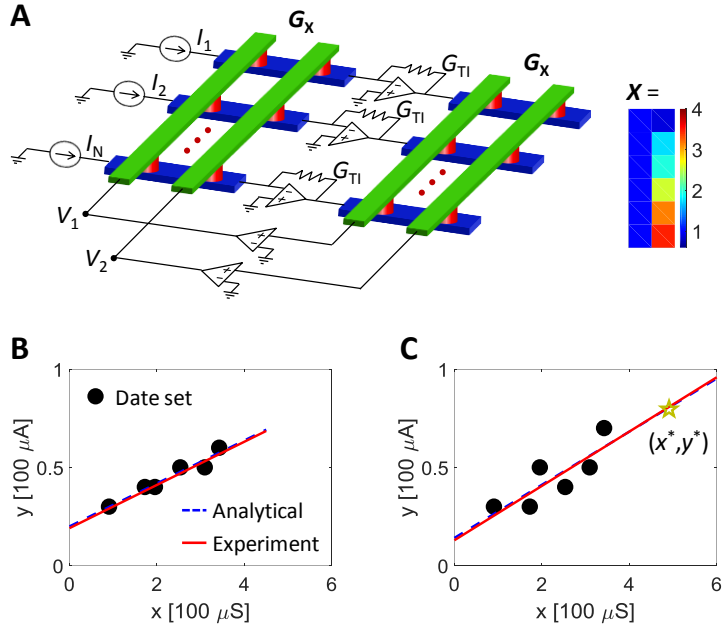

**Fig. 1. Linear regression circuit and experiments.** (**A**) Schematic illustration of the crosspoint circuit for solving linear regression with the pseudoinverse method. The conductance transformation unit is $G_0 = 100$ $\mu$S. The feedback conductance $G_{TI}$ of TIA is equal to $G_0$. A representative matrix $X$ for simple linear regression of 6 data points is also shown. (**B**) Linear regression of a 6-point dataset defined by the second column of $X$ in (**A**) on the *x*-axis and the input currents on the *y*-axis. The figure also shows the analytical and experimental regression lines, the latter being obtained as the measured voltages $v$ in the crosspoint circuit as regression weights. (**C**) A second 6-point regression experiment with the same vector $X$ in (**A**) and a different set of input currents. A new input value $x^* = 4.91$ was stored in an additional line of the left crosspoint array, thus enabling the one-step prediction along a sequence. The measured prediction $y^* = 0.727$ is consistent with experimental and analytical regression lines.



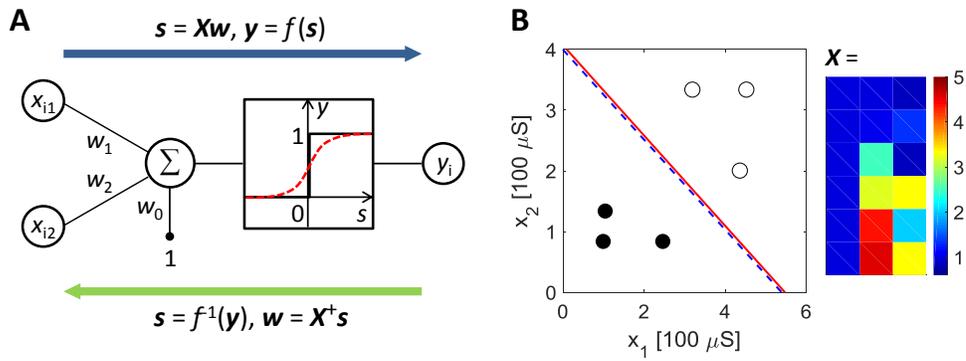

**Fig. 2. Logistic regression experiments.** (**A**) Illustration of a logistic regression model, consisting of the summation of weighted input signals being processed by a nonlinear activation function such as the sigmoid function (dash line) or the step function (solid line). The backward logit transformation is indicated by the bottom arrow. (**B**) A logistic regression of 6 data points divided in 2 classes. The input matrix $X$ is also shown, including a first column of discrete resistors, and a second and third columns storing the independent variables $x_1$ and $x_2$. The regression lines obtained from analytical and experimental results are also shown. The line provides the boundary line for data classification.



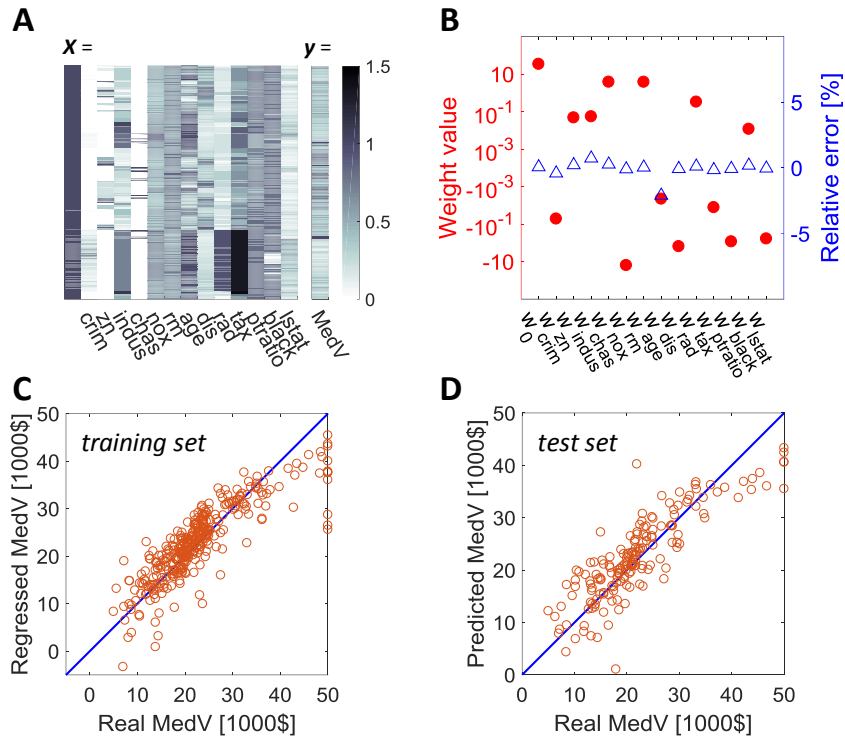

**Fig. 3. Linear regression of the Boston housing dataset.** (**A**) Matrix $X$ including the 13 attributes for the 333 houses in the training set, and input vector $y$ of house prices. The same color bar was used for clarity, while the conductance and current units are assumed equal to 10 $\mu$S and 10 $\mu$A, respectively. (**B**) Calculated weights of the linear regression obtained by simulation of the crosspoint circuit and the relative errors with respect to the analytical results. (**C**) Correlation plot of the regression price of the training samples obtained by the simulated weights, as a function of the real dataset price. The small standard deviation $\sigma_P = 4,733\$$ supports the accuracy of the regression. (**D**) Same as (**C**), but for the test samples. A slightly larger standard deviation $\sigma_P = 4,779\$$ is obtained.

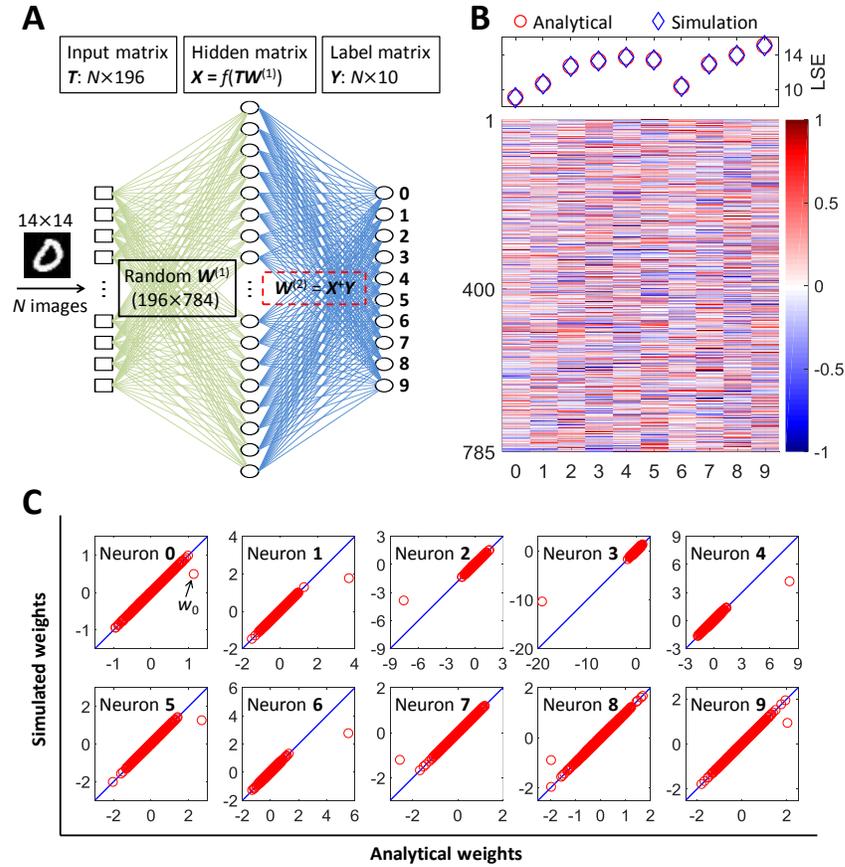

**Fig. 4. Training of a 2-layer neural network for MNIST digit recognition.** (**A**) Illustration of the 2-layer neural network, where the first-layer weights are random, while the second-layer weights are computed by the pseudoinverse method in circuit simulations. (**B**) Color plot of the second-layer weight matrix $W^{(2)}$ obtained by circuit simulations. Each column contains the weights of synapses connected to an output neuron, and was computed in one step by the crosspoint circuit. As a result, only 10 operations were needed to train the network. The LSEs of simulated and analytical weights are also shown for each neuron. (**C**) Correlation plots of the simulated weights as a function of the analytical weights for each of the 10 output neurons. Only the bias weight $w_0$ shows a deviation from the analytical results, though not affecting the recognition accuracy.